\documentclass{article}

\usepackage{aaai18}

\usepackage[utf8]{inputenc} % allow utf-8 input
\usepackage[T1]{fontenc}    % use 8-bit T1 fonts
\usepackage{hyperref}       % hyperlinks
\usepackage{url}            % simple URL typesetting
\usepackage{booktabs}       % professional-quality tables
\usepackage{amsfonts}       % blackboard math symbols
\usepackage{nicefrac}       % compact symbols for 1/2, etc.
\usepackage{microtype}      % microtypography
\usepackage{natbib}
\usepackage{times}
\usepackage{url}
\usepackage{latexsym}
\usepackage[T1]{fontenc}    % use 8-bit T1 fonts
\usepackage{hyperref}       % hyperlinks
\usepackage{url}            % simple URL typesetting
\usepackage{booktabs}       % professional-quality tables
\usepackage{amsfonts}       % blackboard math symbols
\usepackage{nicefrac}       % compact symbols for 1/2, etc.
\usepackage{microtype}      % microtypography
\usepackage{amsmath}
\usepackage{mathtools}
\usepackage{graphicx,color,xcolor}
\usepackage{multirow}
\usepackage{enumitem} 
\usepackage{latexsym}
\usepackage{bm}
\usepackage{algorithm}
\usepackage{wrapfig}
\usepackage{subcaption}
\usepackage[noend]{algpseudocode}

\usepackage{todonotes}

\graphicspath{{./figures}}

\usepackage{lipsum}

\makeatletter
\def\BState{\State\hskip-\ALG@thistlm}
\makeatother
\title{Search Engine Guided Neural Machine Translation}
\def \nyu{$^\ddag$}
\def \hku{$^\dagger$}

\author{{\bf Jiatao Gu\hku, Yong Wang\hku, Kyunghyun Cho\nyu and Victor O.K. Li\hku}
       \\
       \hku The University of Hong Kong \ \ \ \nyu New York University, CIFAR Azrieli Global Scholar \\
       {\tt\hku\{jiataogu, wangyong, vli\}@eee.hku.hk \ \nyu kyunghyun.cho@nyu.edu} 
       }

\date{}

\begin{document}
\maketitle

\begin{abstract}
In this paper, we extend an attention-based neural machine translation (NMT) model by allowing it to access an entire training set of parallel sentence pairs even after training. The proposed approach consists of two stages. In the first stage--retrieval stage--, an off-the-shelf, black-box search engine is used to retrieve a small subset of sentence pairs from a training set given a source sentence. These pairs are further filtered based on a fuzzy matching score based on edit distance. In the second stage--translation stage--, a novel translation model, called search engine guided NMT (SEG-NMT), seamlessly uses both the source sentence and a set of retrieved sentence pairs to perform the translation. Empirical evaluation on three language pairs (En-Fr, En-De, and En-Es) shows that the proposed approach significantly outperforms the baseline approach and the improvement is more significant when more relevant sentence pairs were retrieved. 
\end{abstract}

\section{Introduction}

Neural machine translation is a recently proposed paradigm in machine translation, where a single neural network, often consisting of encoder and decoder recurrent networks, is trained end-to-end to map from a source sentence to its corresponding translation\citep{bahdanau2014neural,cho2014learning,sutskever2014sequence,kalchbrenner2013recurrent}. The success of neural machine translation, which has already been adopted by major industry players in machine translation\citep{wu2016google,crego2016systran}, is often attributed to the advances in building and training recurrent networks as well as the availability of large-scale parallel corpora for machine translation. 

Neural machine translation is most characteristically distinguished from the existing approaches to machine translation, such as phrase-based statistical machine translation\citep{koehn2003statistical}, in that it projects a sequence of discrete source symbols into a continuous space and decodes back the corresponding translation. This allows one to easily incorporate other auxiliary information into the neural machine translation system as long as such auxiliary information could be encoded into a continuous space using a neural network. This property has been noticed recently and used for building more advanced translation systems such as multilingual translation~\citep{firat2016multi,luong2015multi}, multi-source translation~\citep{zoph2016multi,firat2016zero}, multimodal translation~\citep{caglayan2016does} and syntax guided translation~\citep{nadejde2017syntax,eriguchi2017learning}. 

In this paper, we first notice that this ability in incorporating arbitrary meta-data by neural machine translation allows us to naturally extend it to a %non-parametric%
model in which a neural machine translation system explicitly takes into account a full training set consisting of source-target sentence pairs (in this paper we refer them as a general translation memory). We can build a neural machine translation system that considers not only a given source sentence, which is to be translated but also a set of training sentence pairs in the process of translation. To do so, we propose a novel extension of attention-based neural machine translation that seamlessly fuses two information streams, each of which corresponds to the current source sentence and a set of training sentence pairs, respectively. 

A major technical challenge, other than designing such a neural machine translation system, is the scale of a training parallel corpus which often consists of hundreds of thousands to millions of sentence pairs. We address this issue by incorporating an off-the-shelf black-box search engine into the proposed neural machine translation system. The proposed approach first queries a search engine, which indexes a whole training set, with a given source sentence, and the proposed neural translation system translates the source sentence while incorporating all the retrieved training sentence pairs. In this way, the proposed translation system automatically adapts to the search engine and its ability to retrieve relevant sentence pairs from a training corpus.

We evaluate the proposed search engine guided neural machine translation (SEG-NMT) on three language pairs (En-Fr, En-De, and En-Es, in both directions) from JRC-Acquis Corpus\citep{steinberger2006jrc} which consists of documents from a legal domain. This corpus was selected to demonstrate the efficacy of the proposed approach when a training corpus and a set of test sentences are both from a similar domain. Our experiments reveal that the proposed approach exploits the availability of the retrieved training sentence pairs very well, achieving significant improvement over the strong baseline of attention-based neural machine translation\citep{bahdanau2014neural}.

\section{Background}

\subsection{Neural Machine Translation}

In this paper, we start from a recently proposed, and widely used, attention-based neural machine translation model\citep{bahdanau2014neural}. The attention-based neural translation model is a conditional recurrent language model of a conditional distribution $p(Y|X)$ over all possible translations $Y=\left\{ y_1, \ldots, y_T \right\}$ given a source sentence $X=\left\{ x_1, \ldots, x_{T_x}\right\}$. This conditional recurrent language model is an autoregressive model that estimates the conditional probability as
$p(Y|X) = \prod_{t=1}^T p(y_t | y_{<t}, X)$. Each term on the right hand side is approximated by a recurrent network by 
\begin{align}
\label{eq.prob}
p(y_t|y_{<t}, X)\propto \exp\left(g\left(y_t, z_t; \theta_g\right)\right),
\end{align}
where 
$z_t = f(z_{t-1}, y_{t-1}, c_t(X, z_{t-1}, y_{t-1}); \theta_f)$. $g$ and $f$ correspond to a read-out function that maps the hidden state $z_t$ into a distribution over a target vocabulary, and a recurrent activation function that summarizes all the previously decoded target symbols $y_1, \ldots, y_{t-1}$ with respect to the time-dependent context vector $c_t(X; \theta_e)$, respectively. Both of these functions are parametrized, and their parameters are learned jointly to maximize the log-likelihood of a training parallel corpus.
$c_t(X, z_{t-1}, y_{t-1})$ is composed of a bidirectional recurrent network encoder and an attention mechanism. The source sequence $X$ is first encoded into a set of annotation vector $\left\{ h_1, \ldots, h_{T_x} \right\}$, each of which is a concatenation of the hidden states of the forward and reverse recurrent networks. The attention mechanism, which is implemented as a feedforward network with a single hidden layer, then computes an attention score $\alpha_{t, \tau}$ for each hidden state $h_{\tau}$ given the previously decoded target symbol $y_{t-1}$ and the previous decoder hidden state $z_{t-1}$:
\begin{align*}
\alpha_{t, \tau} = \frac{\exp\left\{ \phi_{\text{att}} (h_{\tau}, y_{t-1}, z_{t-1}) \right\}}
{\sum_{\tau'=1}^{T_x} \exp\left\{ \phi_{\text{att}} (h_{\tau'}, y_{t-1}, z_{t-1}) \right\}}.
\end{align*}
These attention scores are used to compute the time-dependent context vector $c_t$ as 
\begin{align}
\label{eq.context}
c_t = \sum_{\tau=1}^{T_x} \alpha_{t, \tau} h_{\tau}.
\end{align}

The attention-based neural machine translation system is end-to-end trained to maximize the likelihood of a correct translation given a corresponding source sentence. During testing, a given source sentence is translated by searching for the most likely translation from a trained model. The entire process of training and testing can be considered as compressing the whole training corpus into a neural machine translation system, as the training corpus is discarded once training is over. 

\subsection{Translation Memory}

Translation memory is a computer-aided translation tool widely used by professional human translators. It is a database of pairs of source phrase and its translation. This database is constructed incrementally as a human translator translates sentences. When a new source sentence is present, a set of (overlapping) phrases from the original sentence are queried against the translation memory, and the corresponding entries are displayed to the human translator to speed up the process of translation. Due to the problem of sparsity (Sec.5.2 of \cite{cho2015natural} ), exact matches rarely occur, and approximate string matching is often used.

In this paper, we consider a more general notion of translation memory in which not only translation phrase pairs but any kind of translation pairs are stored. In this more general definition, a training parallel corpus is also considered a translation memory. This saves us from building a phrase table\citep{koehn2003statistical}, which is yet another active research topic, but requires us to be efficient and flexible in retrieving relevant translation pairs given a source sentence, as the issue of data sparsity amplifies. This motivates us to come up with an efficient query algorithm tied together with a downstream translation model that can overcome the problem of data sparsity.

\begin{figure*}[t]
%\label{framework}
\begin{minipage}{0.82\textwidth}
% \vspace{-10pt}
\centering
\includegraphics[width=\linewidth]{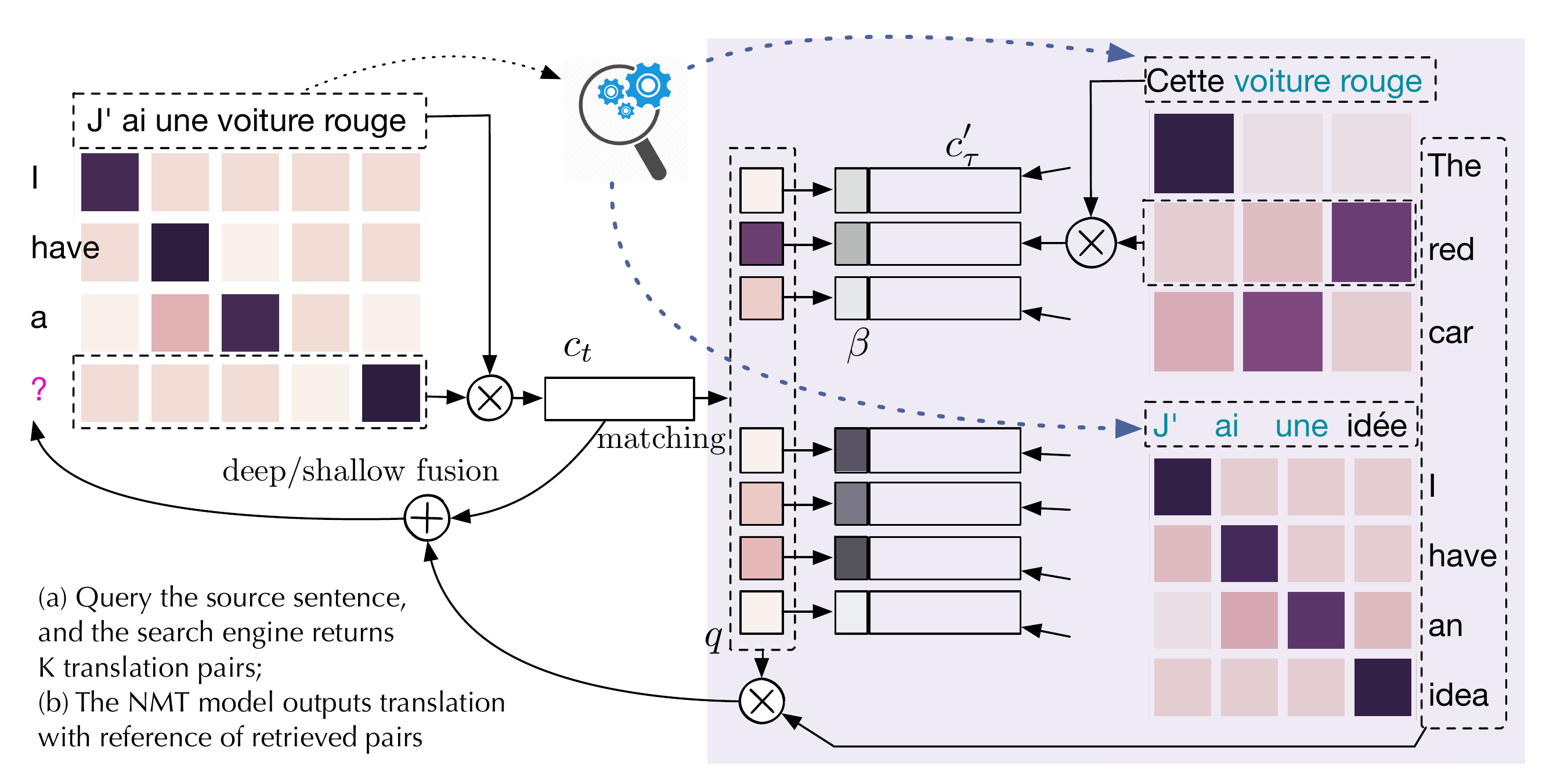}
\end{minipage}
\hfill
\begin{minipage}{0.17\textwidth} 
\caption{\label{fig.tmnmt} The overall architecture of the proposed SEG-NMT. The shaded box includes the module which handles a set of translation pairs retrieved in the first stage. The heat maps represent the attention scores between the source sentences (left-to-right) and the corresponding translations (top-to-down).}
\end{minipage}
% \vspace{-4mm}
\end{figure*}

\section{Search Engine Guided Non-Parametric Neural Machine Translation}

%\vspace{-3pt}
We propose a non-parametric
neural machine translation model guided by an off-the-shelf, efficient search engine. Unlike the conventional neural machine translation system, the proposed model does not discard a training corpus but maintain and actively exploit it in the test time. This effectively makes the proposed neural translation model a fully non-parametric model.

The proposed nonparametric
neural translation model consists of two stages. The first stage is a retrieval stage, in which the proposed model queries a training corpus, or equivalently a translation memory, to retrieve a set of source-translation pairs given a current source sentence. To maximize the computational efficiency, first we utilize an off-the-shelf, highly-optimized search engine to quickly retrieve a large set of similar source sentences, and their translations, after which the top-$K$ pairs are selected using approximate string matching based on edit distance. 

In the second stage, a given source sentence is translated by an attention-based neural machine translation model, which we refer to as a {\it search engine guided neural machine translation} (SEG-NMT), and incorporates the retrieved translation pairs from the first stage. In order to maximize the use of the retrieved pairs, we build a novel extension of the attention-based model that performs attention not only over the source symbols but also over the retrieved symbols (and their respective translations). We further allow the model an option to copy over a target symbol directly from the retrieved translation pairs. The overall architecture with a simple translation example of the proposed SEG-NMT is shown in Fig.~\ref{fig.tmnmt} for reference.

% \begin{wrapfigure}{R}{0.5\textwidth}
% \vspace{-4mm}
% \begin{minipage}{0.5\textwidth}
% \end{minipage}
% \vspace{-3mm}
% \end{wrapfigure}

%\vspace{-3pt}
\subsection{Retrieval Stage}

We refer to the first stage as a {\it retrieval stage}. In this stage, we go over the entire training set $\mathcal{M}=\left\{ (X^n, Y^n)\right\}_{n=1}^N$ to find pairs whose source side is similar to a current source $X$. That is, we define a similarity function $s(X, X')$, and find $(X^n, Y^n)$ where $s(X, X^n)$ is large. 

%\vspace{-7pt}
\paragraph{Similarity score function $s$}

In this paper, we constrain ourselves to a setting in which only a neural translation model is trainable. That is, we do not assume the availability of other trainable sentence similarity functions. This allows us to focus entirely on the effectiveness of the proposed algorithm while being agnostic to the choice of similarity metric. Under this constraint, we follow an earlier work by \citep{li2016phrase} and use a fuzzy matching score which is defined as 
\begin{align}
\label{eq.fuzzy}
s_{\text{fuzzy}}(X, X') = 1 - \frac{D_{\text{edit}}(X, X')}{\max\left(|X|, |X'|\right)},
\end{align}
where $D_{\text{edit}}$ is an edit distance. 

\begin{algorithm}[htpb]
\caption{Greedy selection procedure to maximize the coverage of the source symbols.} %\alert{KC: put details of each symbol inside the algorithm.}}
\label{algo1}
\begin{algorithmic}[1]
\small
\Require{input $X$, translation memory $\mathcal{M}$}
\State Obtain the subset $\tilde{M}\subseteq M$ using an off-the-shelf search engine;
\State Re-rank retrieved pairs $\left(X', Y' \right) \in \tilde{M}$ using the similarity score function $s$ in descending order;
\State Initialize the dictionary of selected pairs $R = \emptyset$; 
\State Initialize the coverage score $c=0$;
\For{$k=1...|\tilde{M}|$}  
\State $c_{\text{tmp}}= \sum_{x \in X}\delta\left[ x \in R.\text{keys} \cup \{X'_k\}\right]/|X|$
\If{$c_{\text{tmp}}> c$} 
\State $c = c_{\text{tmp}}$; $R \leftarrow \{X'_k: Y'_k\}$
\EndIf
\EndFor
\State \textbf{return} $R$
\end{algorithmic}
\end{algorithm}

\vspace{-7pt}
\paragraph{Off-the-shelf Search Engine}

The computational complexity of the similarity search grows linearly with the size of the translation memory which in our case contains all the pairs from a training corpus. Despite the simplicity and computational efficiency of the similarity score in Eq.~\eqref{eq.fuzzy}, this is clearly not practical, as the size of the training corpus is often in the order of hundreds of thousands or even tens of millions. We overcome this issue of scalability by incorporating an off-the-shelf search engine, more specifically Apache Lucene.\footnote{
\url{https://lucene.apache.org/core/}
} 
We then use Lucene to retrieve an initial set of translation pairs based on the source side, and use the similarity score above to re-rank them.

\vspace{-7pt}
\paragraph{Final selection process}
Let $\tilde{\mathcal{M}} \in \mathcal{M}$ be an initial set of translation pairs returned by Lucene. We rank the translation pairs within this set by $s(X, X')$. We design and test two methods for selecting the final set from this initial set based on the similarity scores. The first method is a top-$K$ retrieval, where we simply return the $K$ most similar translation pairs from $\tilde{\mathcal{M}}$. The second method returns an adaptive number of translation pairs based on the coverage of the symbols $x$ in the current source sentence $X$ within the retrieved translation pairs. We select greedily starting from the most similar translation pair, as described in Alg.~\ref{algo1}.

\subsection{Translation Stage} 
In the second stage, we build a novel extension of the attention-based neural machine translation, SEG-NMT, that seamlessly fuses both a current source sentence and a set $\hat{M}$ of retrieved translation pairs. In a high level, the proposed SEG-NMT first stores each target symbol of each retrieved translation pair into a key-value memory\citep{miller2016key}. At each time step of the decoder, SEG-NMT first performs attention over the current source sentence to compute the time-dependent context vector based on which the key-value memory is queried. SEG-NMT fuses information from both context vector of the current source sentence and the retrieved value from the key-value memory to generate a next symbol. 

\vspace{-7pt}
\paragraph{Key-Value Memory}

For each retrieved translation pair $(X', Y') \in \hat{\mathcal{M}}$, we run a full attention-based neural machine translation model,\footnote{
We use a single copy of attention-based model for both key extraction and translation.
}
specified by a parameter set $\theta$, and obtain, for each target symbol $y'_t \in Y'$, a decoder's hidden state $z'_t$ and an associated time-dependent context vector $c'_t$ (see Eq.~\eqref{eq.context} which summarizes a subset of the source sentence $X'$ that best describes $y'_t$). We consider $c'_t$ as a key and $(z'_t, y'_t)$ as a value, and store all of them from all the retrieved translation pairs in a key-value memory. Note that this approach is agnostic to how many translation pairs were retrieved during the first stage.

\vspace{-7pt}
\paragraph{Matching and Retrieval}

At each time step of the SEG-NMT decoder, we first compute the context vector $c_t$ given the previous decoder hidden state $z_t$, the previously decoded symbol $y_{t-1}$ and all the annotation vector $h_{\tau}$'s, as in Eq.~\eqref{eq.context}. This context vector is used as a key for querying the key-value memory. Instead of hard matching, we propose {\it soft matching} based on a bilinear function, where we compute the matching score of each key-value slot by
\begin{align}
\label{eq.score}
    q_{t, \tau}  = \frac{\exp\{E(c_t, c'_\tau)\}}{\sum_{\tau'} \exp\{E(c_t, c'_{\tau'})\}}.
\end{align}
where $E(c_t, c'_\tau) = c_t^TMc'_\tau$ and $M$ is a trainable matrix. 

These scores are used to retrieve a value from the key-value memory. In the case of the decoder's hidden states, we retrieve a weighted sum:
$\tilde{z}_t = \sum_{\tau} q_{t, \tau} z'_{\tau}$;
In the case of target symbols, we consider each computed score as a probability of the corresponding target symbol. That is,
$p_{\text{copy}}(y_{\tau}') = q_{t, \tau},$ similarly to the pointer network\citep{vinyals2015pointer}.

% \begin{wrapfigure}{R}{0.54\textwidth}
% \vspace{-22pt}
% \begin{minipage}{0.54\textwidth}
% \end{minipage}
% \vspace{-5pt}
% \end{wrapfigure}

\vspace{-7pt}
\paragraph{Incorporation}

We consider two separate approaches to incorporating the retrieved values from the key-value memory, motivated by \citep{Gulcehre-Orhan-et-al-2015}. The first approach, called {\it deep fusion}, weighted-average the retrieved hidden state $\tilde{z}_t$ and the decoder's hidden state $z_t$:
\begin{align}
\label{eq.deep}
z_{\text{fusion}} = &\zeta_t \cdot \tilde{z}_t + (1 -\zeta_t) \cdot z_t 
\end{align}
when computing the output distribution $p(y_t|y_{<t},X, \mathcal{M})$ (see Eq.~\eqref{eq.prob}). The second approach is called {\it shallow fusion} and computes the output distribution as a mixture:
\begin{align}
\label{eq.shallow}
\begin{split}
p(y_t| y_{<t}, X, \mathcal{M}) = \zeta_t &p_{\text{copy}}(y_t) \\
+ (1-\zeta_t) &p(y_t|y_{<t}, X).
\end{split}
\end{align}
This is equivalent to copying over a retrieved target symbol $y_{\tau}'$ with the probability of $\zeta_t p_{\text{copy}}(y_{\tau})$ as the next target symbol\citep{gulcehre2016pointing,gu2016incorporating}.% We can build a SEG-NMT by using either or both of these approaches.

In both of the approaches, there is a gating variable $\zeta_t$. As each target symbol may require a different source of information, we let this variable be determined automatically by the proposed SEG-NMT. That is, we introduce another feedforward network that computes $\zeta_t = f_{\text{gate}}(c_t, z_t, \tilde{z}_t)$. This gate closes when the retrieved pairs are not useful for predicting the next target symbol $y_t$, and opens otherwise. 

\begin{algorithm}[H]
\caption{Learning for SEG-NMT}% \alert{TODO: if there's not enough space, we should move this to a supplementary material.}}
\label{algo2}
\begin{algorithmic}[1]
\small
\Require{Search engine  $F_{SS}$, MT model ${\theta}$, SEG model ${\theta'}$, $M, \lambda, \eta$, parallel training set $\mathcal{D}$,  translation memory $\mathcal{M}$.}
\State Initialize $\phi = \{\theta, \theta', M, \lambda, \eta\}$;
\State Set the number of returned answers as $K$;
\While{stopping criterion is not met}
\State Draw a translation pair: $(X, Y)\sim \mathcal{D}$;
\State Obtain memory pairs $\{X'_k, Y'_k\}_{k=1}^K = F_{SS}(X, \mathcal{M})$
\State Reference Memory $C=\emptyset$.
\For{$k=1...K$}   \# generate dynamic keys
\State Let $Y'_k = \{y'_1, ..., y'_{T'}\}, X'_k = \{x'_1, ..., x'_{T'_s}\}$
\For{$\tau=1...T'$}
\State Generate key $c'_\tau = f_{\text{att}}(y'_{<\tau}, X'_k)$
\State Initialize coverage $\beta_{\tau} = 0.$
\State $ C \leftarrow (c'_\tau, y'_\tau, \beta_{\tau})$
\EndFor 
\EndFor
\State Let $Y = \{y_1, ..., y_{T}\}, X = \{x_1, ..., x_{T_s}\}$
\For{$t=1...T$} \# translate each word
\State Generate query $c_t = f_{\text{att}}(y_{<t}, X)$
\For{$\tau=1...T'$}  Read $c'_\tau, y'_\tau,  \beta_{\tau} \in C$
\State Compute the score $q_{t, \tau}$ using Eq.~\ref{eq.match};
\State Gompute the gate $\zeta_t$ with $f_{\text{gate}}$;
\State Update $\beta_{\tau} \leftarrow \beta_{\tau} + q_{t, \tau}\cdot\zeta_t$;
\EndFor
\State Compute the probability $p(y_t|\cdot)$
\State \hspace{10pt} --option1: shallow-fusion, Eq.~\ref{eq.shallow}
\State  \hspace{10pt} --option2: deep-fusion, Eq.~\ref{eq.deep}
\EndFor 
\State Update $\phi \leftarrow \phi  +\gamma\frac{\partial}{\partial \phi} \sum_{t=1}^T\log p(y_t|\cdot)$
\EndWhile
\end{algorithmic}
\end{algorithm}

\paragraph{Coverage}

In the preliminary experiments, we notice that the access pattern of the key-value memory was highly skewed toward only a small number of slots. Motivated by the coverage penalty from \citep{tu2016modeling}, we propose to augment the bilinear matching function (in Eq.~\eqref{eq.score}) with a coverage vector $\beta_{t,\tau}$ such that
\vspace{-3pt}
\begin{align}
\label{eq.match}
     E(c_t, c'_\tau) = c_t^TMc'_\tau - \lambda \beta_{t-1, \tau},
\vspace{-3pt}
\end{align}
where the coverage vector is defined as 
$    \beta_{t, \tau}= \sum_{t'=1}^t q_{t', \tau}\cdot\zeta_{t'}$. $\lambda$ is a trainable parameter.

\subsection{Learning and Inference}

The proposed model, including both the first and second stages, can be trained end-to-end to maximize the log-likelihood given a parallel corpus. For practical training, we preprocess a training parallel corpus by augmenting each sentence pair with a set of translation pairs retrieved by a search engine, while ensuring that the exact copy is not included in the retrieved set. See Alg.~\ref{algo2} for a detailed description. During testing, we search through the whole training set to retrieve relevant translation pairs. Similarly to a standard neural translation model, we use beam search to decode the best translation given a source sentence.

\section{Related Work}

The principal idea of SEG-NMT shares major similarities with the example-based machine translation (EBMT)~\citep{Zhang2005AnEP,callison2005scaling,phillips2012modeling} which indexes parallel corpora with suffix arrays and retrieves translations on the fly at test time. However, to the best of our knowledge, SEG-NMT is the first work incorporating any attention-based neural machine translation architectures and can be trained end-to-end efficiently, showing superior performance and scalability compared to the conventional statistical EBMT.

% \paragraph{Multilingual neural machine translation}
%\vspace{-2mm}
SEG-NMT has also been largely motivated by recently proposed multilingual attention-based neural machine translation models\citep{firat2016multi,zoph2016multi}. Similar to these multilingual models, our model takes into account more information than a current source sentence. This allows the model to better cope with any uncertainty or ambiguity arising from a single source sentence. More recently, this kind of larger context translation has been applied to cross-sentential modeling, where the translation of a current sentence is done with respect to previous sentences\citep{jean2017does,wang2017exploiting}. 

% \paragraph{Nearest-neighbour image caption generation}

\cite{devlin2015exploring} proposed an automatic image caption generation model based on nearest neighbours. In their approach, a given image is queried against a set of training pairs of images and their corresponding captions. They then proposed to use a median caption among those nearest neighboring captions, as a generated caption of the given image. This approach shares some similarity with the first stage of the proposed SEG-NMT. However, unlike their approach, we {\it learn to generate} a sentence rather than simply choose one among retrieved ones. 

% \paragraph{Large-scale question-answering}

\cite{bordes2015large} proposed a memory network for large-scale simple question-answering using an entire Freebase\citep{bollacker2008freebase}. The output module of the memory network used simple $n$-gram matching to create a small set of candidate facts from the Freebase. Each of these candidates was scored by the memory network to create a representation used by the response module. This is similar to our approach in that it exploits a black-box search module ($n$-gram matching) for generating a small candidate set. 

% \paragraph{Neural episodic control}

%The proposed method makes the attention-based neural machine translation model non-parametric by incorporating a key-value memory that stores a training set.%
A similar approach was very recently proposed for deep reinforcement learning by \cite{pritzel2017neural}, where they store pairs of observed state and the corresponding (estimated) value in a key-value memory to build a non-parametric deep Q network. We consider it as a confirmation of the general applicability of the proposed approach to a wider array of problems in machine learning.
In the context of neural machine translation, \cite{kaiser2017learning} also proposed to use an external key-value memory to remember training examples in the test time. Due to the lack of efficient search mechanism, they do not update the memory jointly with the translation model, unlike the proposed approach in this paper.

One important property of the proposed SEG-NMT is that it relies on an external, black-box search engine to retrieve relevant translation pairs. Such a search engine is used both during training and testing, and an obvious next step is to allow the proposed SEG-NMT to more intelligently query the search engine, for instance, by reformulating a given source sentence. Recently, \cite{nogueira2017task} proposed task-oriented query reformulation in which a neural network is trained to use a black-box search engine to maximize the recall of relevant documents, which can be integrated into the proposed SEG-NMT. We leave this as future work.

%\vspace{-3pt}
\section{Experimental Settings}
%\vspace{-3pt}
\paragraph{Data}

% \begin{wraptable}{R}{0.5\textwidth} 
% \vspace{-4mm}
\begin{table}
\small
\centering
\begin{tabular}{l|r|r|r}
\toprule
 Dataset& En-Fr & En-De  & En-Es\\
\midrule
\# Train Pairs &744,528 &717,096 &697,187 \\
\# Dev Pairs &2,665 &2,454 &2,533  \\
\# Test Pairs &2,655 &2,483 &2,596  \\
\midrule
\# En/sent. &29.44&33.43 &32.10  \\
\# Other/sent. &33.34 &33.44 &34.95  \\
\bottomrule
\end{tabular} \vspace{-2pt}
\caption{\label{table.dataset} Statistics from the JRC-Acquis corpus. We use BPE subword symbols.}
\vspace{-4mm}
\end{table}

%\end{wraptable}   

We use the JRC-Acquis corpus\citep{steinberger2006jrc} for evaluating the proposed SEG-NMT model.\footnote{
\url{http://optima.jrc.it/Acquis/JRC-Acquis.3.0/corpus/}
} The JRC-Acquis corpus consists of the total body of European Union (EU) law applicable to the member states. The text in this corpus is well structured, and most of the text in this corpus are related, making it an ideal test bed to evaluate the proposed SEG-NMT which relies on the availability of appropriate translation pairs from a training set. This corpus was also used by \citep{li2016phrase} in investigating the combination of translation memory and phrase-based statistical machine translation, making it suitable for our proposed method to evaluate on.

We select three language pairs, namely, En-Fr, En-Es, and En-De, for evaluation. For each language pair, we uniformly select 3000 sentence pairs at random for both the development and test sets. The rest is used as a training set, after removing any sentence which contains special characters only. We use sentences of lengths up to 80 and 100 from the training and dev/test sets respectively. We do not lowercase the text, and use byte-pair encoding (BPE)\citep{sennrich2015neural} to extract a vocabulary of 20,000 subword symbols. See Table~\ref{table.dataset} for detailed statistics.

%\vspace{-9pt}
\paragraph{Retrieval Stage}

We use Apache Lucene to index a whole training set and retrieve 100 pairs per source sentence for the initial retrieval. These 100 pairs are scored against the current source sentence using the fuzzy matching score from Eq.~\eqref{eq.fuzzy} to select top-$K$ relevant translation pairs. We vary $K$ among $1$ and $2$ during training and among $1, 2, 4, 8, 16$ during testing to investigate the trade-off between retrieval and translation quality. During testing, we also evaluate the effect of adaptively deciding the number of retrieved pairs using the proposed greedy selection algorithm (Alg.~\ref{algo1}). 

%\vspace{-9pt}
\paragraph{Translation Stage}

We use a standard attention-based neural machine translation model\citep{bahdanau2014neural} with 1,024 gated recurrent units(GRU)\citep{cho2014learning} on each of the encoder and decoder. We train both the vanilla model as well as the proposed SEG-NMT based on this configuration from scratch using Adam\citep{kingma2014adam} with the initial learning rate set to 0.001. We use a minibatch of up to 32 sentence pairs. We early-stop based on the development set performance. For evaluation, we use beam search with width set to 5.

In the case of the proposed SEG-NMT, we parametrize the metric matrix $M$ in the similarity score function from Eq.~\eqref{eq.match} to be diagonal and initialized to an identity matrix. $\lambda$ in Eq.~\eqref{eq.match} is initialized to $0$. The gating network $f_{\text{gate}}$ is a feedforward network with a single hidden layer, just like the attention mechanism $f_{\text{att}}$. We use either deep fusion or shallow fusion in our experiments.

\section{Result and Analysis}
%\vspace{-3pt}
% \begin{wraptable}{R}{0.61\textwidth} 
\begin{table}
\small
%\vspace{-4mm}
  \centering
    \begin{tabular}{l|l|cc|cc|cc}
    \toprule
    \multicolumn{2}{c|}{\multirow{2}[2]{*}{}} & \multicolumn{2}{c|}{En-Fr} & \multicolumn{2}{c|}{En-De} & \multicolumn{2}{c}{En-Es} \\
    \multicolumn{2}{c|}{} & \multicolumn{1}{c}{$\rightarrow$} & \multicolumn{1}{c|}{$\leftarrow$}  & \multicolumn{1}{c}{$\rightarrow$} & \multicolumn{1}{c|}{$\leftarrow$} & \multicolumn{1}{c}{$\rightarrow$} & \multicolumn{1}{c}{$\leftarrow$} \\
    \midrule
    \multirow{3}[2]{*}{\rotatebox{90}{Dev}} 
         & TM    & \multicolumn{1}{c}{46.62} & 42.53 & \multicolumn{1}{c}{34.99} & \multicolumn{1}{c|}{42.45} & \multicolumn{1}{c}{40.84} & \multicolumn{1}{c}{39.71} \\
         & NMT   & \multicolumn{1}{c}{58.95} & 59.69 & \multicolumn{1}{c}{44.94} & \multicolumn{1}{c|}{50.20} & \multicolumn{1}{c}{50.54} & \multicolumn{1}{c}{55.02} \\
         & Copy & \multicolumn{1}{c}{60.34} & 61.61 & - & - & - & -\\
         & Ours & \multicolumn{1}{c}{\textbf{64.16}} & \textbf{64.64} & \multicolumn{1}{c}{\textbf{49.26}} & \multicolumn{1}{c|}{\textbf{55.63}} & \multicolumn{1}{c}{\textbf{57.62}} & \multicolumn{1}{c}{\textbf{60.28}} \\
    \midrule
    \multirow{3}[2]{*}{\rotatebox{90}{Test}}
             & TM    & \multicolumn{1}{c}{46.64} & 43.17 & \multicolumn{1}{c}{34.61} & \multicolumn{1}{c|}{41.83} & \multicolumn{1}{c}{39.55} & \multicolumn{1}{c}{37.73} \\
          & NMT   &   59.42    &  60.11     &   43.98    &   49.74   &   50.48    &  54.66\\
          & Copy & \multicolumn{1}{c}{60.55} & 62.02 & - & - & - & -\\
          & Ours & \textbf{64.60} & \textbf{65.11} & \textbf{48.80}  &  \textbf{55.33}  &  \textbf{57.27}  &  \textbf{59.34} \\
    \bottomrule
    \end{tabular}%
     \caption{The BLEU scores on JRC-Acquis corpus.}
     %\vspace{-3mm}
  \label{tab:bleu}%
%\end{wraptable}
\end{table}

In Table~\ref{tab:bleu}, we present the BLEU scores obtained on all the three language pairs (both directions each) using three approaches; TM -- a carbon copy of the target side of a retrieved translation pair with the highest matching score, NMT - a baseline translation model, and our proposed SEG-NMT model. It is evident from the table that the proposed SEG-NMT significantly outperforms the baseline model in all the cases, and that this improvement is not merely due to their copying over the most similar translation from a training set. For Fr-En and En-Fr, we also present the performance of using a ``CopyNet''~\citep{gu2016incorporating} variant which uses a copying mechanism directly over the target side of the searched translation pair. This CopyNet variant helps but not as much as the proposed approach. We conjecture this happens because our proposal of using a key-value memory captures the relationship between the source and target tokens in the retrieved pairs more tightly.

\begin{figure}
\centering
\includegraphics[width=\linewidth,clip=True,trim=0 15 0 30]{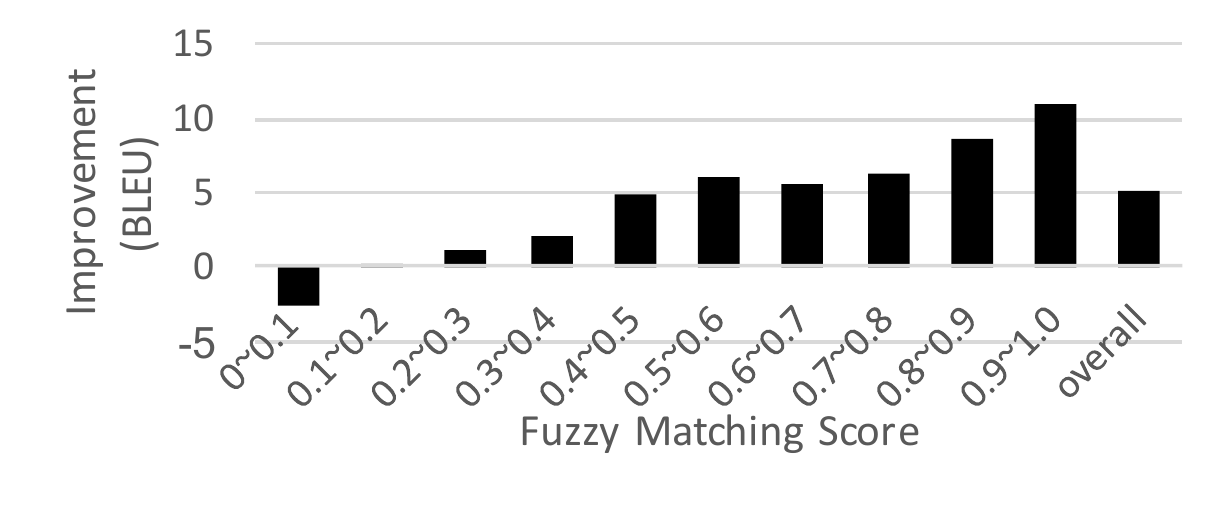}
\vspace{-3mm}
\caption{
\label{fig:fuzzy_improv}
The improvement over the baseline by SEG-NMT on Fr$\to$En w.r.t. the fuzzy matching scores of one retrieved translation pair. 
}
\vspace{-4mm}
\end{figure}

\vspace{-7pt}
\paragraph{Fuzzy matching score v.s. Quality}
%\begin{wrapfigure}{L}{0.5\textwidth}
%\end{wrapfigure}
For Fr$\to$En, we broke down the development set into a set of bins according to the matching score of a retrieved translation pair, and computed the BLEU score for each bin. As shown in Fig.~\ref{fig:fuzzy_improv}, we note that the improvement grows as the relevance of the retrieved translation pair increases. This verifies that SEG-NMT effectively exploits retrieved translation pairs, but also suggests a future improvement for the case in which no relevant translation pair exists in a training set. 
%\begin{wrapfigure}{L}{0.5\textwidth}
\begin{figure}
%\vspace{-5pt}
\centering
\includegraphics[width=\linewidth,clip=True,trim=0 5 0 20]{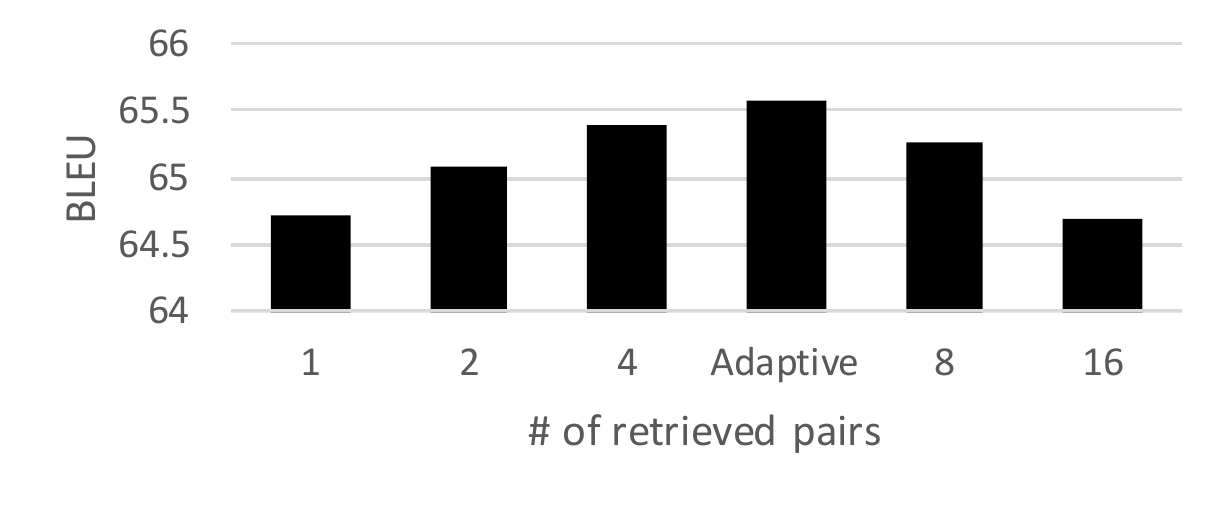}
\vspace{-6mm}
\caption{
\label{fig:bleu_retrieved}
The BLEU scores on Fr$\to$En using varying numbers of retrieved translation pairs during testing. The model was trained once. ``Adaptive'' refers to the proposed greedy selection in Alg.~\ref{algo1}.
}
\vspace{-4mm}
\end{figure}

\begin{figure*}[t]
% \centering
% \includegraphics[width=\linewidth]{figures/example-a.pdf}
% %\vspace{-5pt}
% %\caption{\label{fig.example} Examples}-
% \centering
% \includegraphics[width=\linewidth]{figures/example-b.pdf}
% %\vspace{mm}
%\begin{minipage}{0.82\textwidth}
\centering
\includegraphics[width=0.8\linewidth]{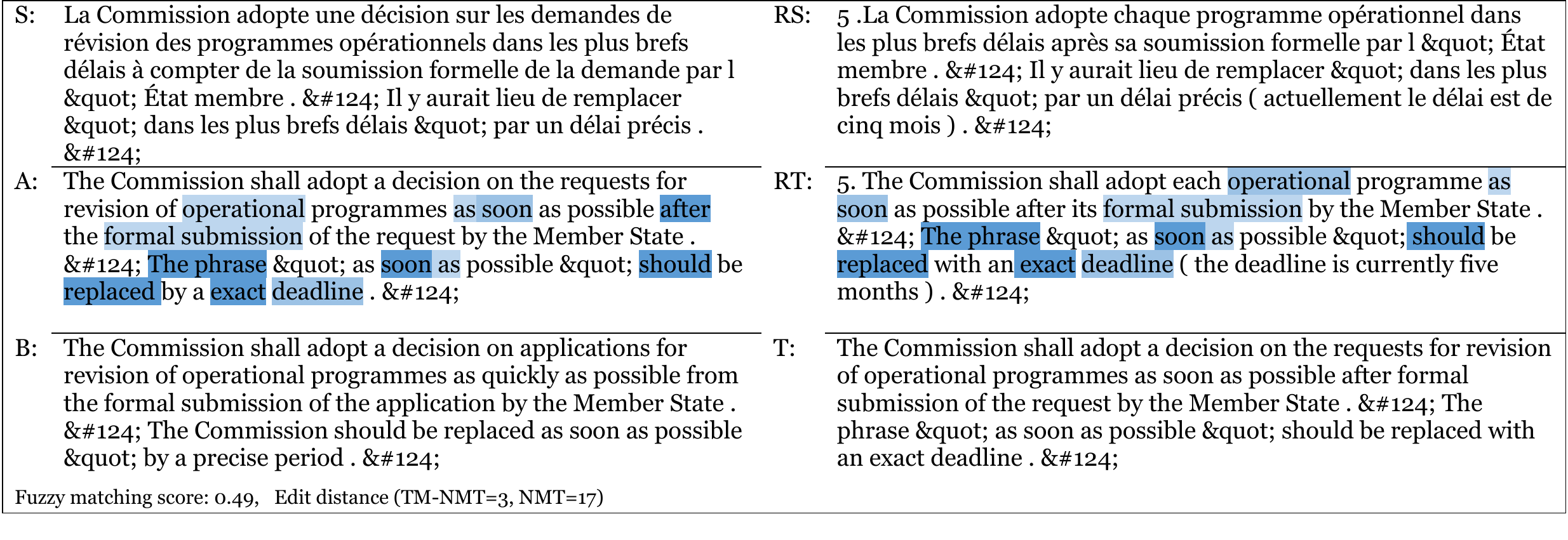}
%\vspace{-1mm}
\centering
\includegraphics[width=0.8\linewidth]{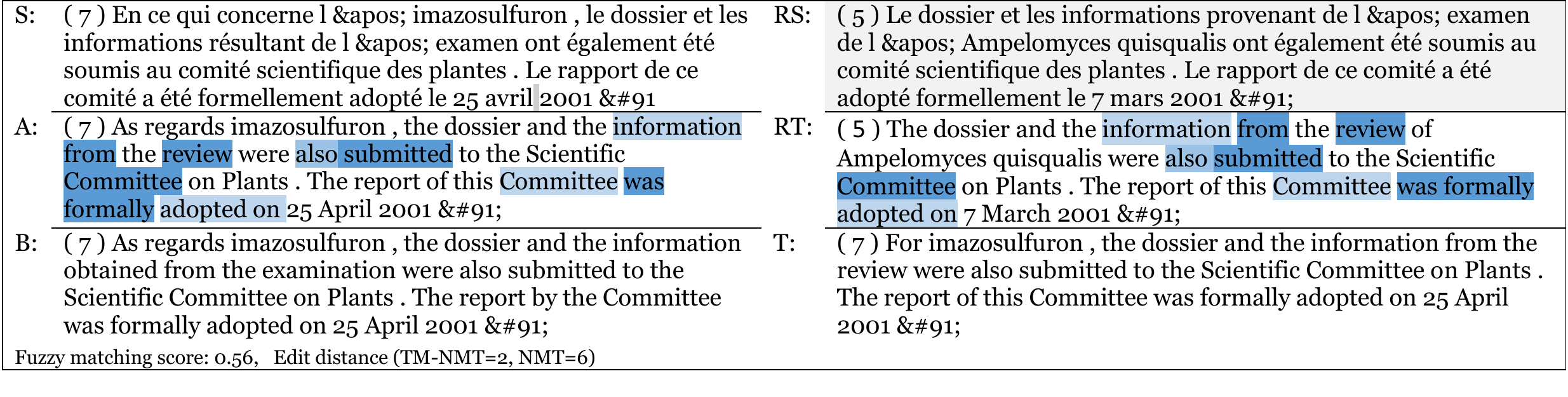}
%\vspace{-1mm}
\centering
\includegraphics[width=0.8\linewidth]{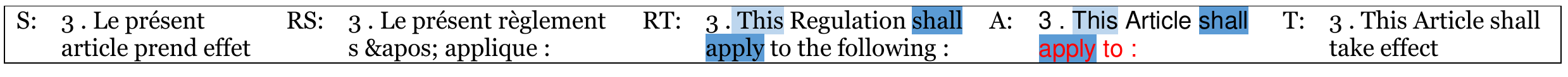}
%\end{minipage}
%\caption{\label{fig.example} Examples}
%\vspace{0mm}
\caption{\label{fig:examples} 
Three examples from the Fr$\to$En test set. For the proposed SEG-NMT model, one translation pair is retrieved from the training set. Each token in the translation by the proposed approach and its corresponded token (if it exists) in the retrieved pair are shaded in blue according to the gating variable $\zeta_t$ from Eq.~\eqref{eq.shallow}. In all, we show: (S) the source sentence. (RS) the source side of a retrieved pair. (RT) the target side of the retrieved pair.
(A) the translation by the proposed approach. (B) the translation by the baseline. (T) the reference translation.
}

%\vspace{-5mm}
\end{figure*}

\vspace{-7pt}
\paragraph{Effect of the \# of Retrieved Translation Pairs}
Once the proposed model is trained, it can be used with a varying number of retrieved translation pairs. We test the model trained on Fr$\to$En with different numbers of retrieved translation pairs, and present the BLEU scores in Fig.~\ref{fig:bleu_retrieved}. We notice that the translation quality increases as the number of retrieved pairs increase up to approximately four, but from there on it degrades. We believe this happens as the retrieved sentences become less related to the current source sentence. The best quality was achieved when the proposed greedy selection algorithm in Alg.~\ref{algo1} was used, in which case 4.814 translation pairs were retrieved on average. 

\vspace{-7pt}
\paragraph{Deep vs. Shallow Fusion}
On both directions of En-Fr, we implemented and tested both deep and shallow fusion (Eqs.~\eqref{eq.deep}--\eqref{eq.shallow}) for incorporating the information from the retrieved translation pairs. With deep fusion only, the BLEU scores on the development set improved over the baseline by 1.30 and 1.20 respectively, while the improvements were 5.21 and 4.95, respectively. This suggests that the proposed model effectively exploits the availability of target symbols in the retrieved translation pairs. All other experiments were thus done using shallow fusion only.

\vspace{-7pt}
\paragraph{Examples}
We list two good examples and one in which the proposed method makes a mistake, in Fig.~\ref{fig:examples}. From these examples, we see that the proposed SEG-NMT selects a term or phrase used in a retrieved pair whenever there are ambiguities or multiple correct translations. For instance, in the first example, SEG-NMT translated ``pr\'ecis'' into ``exact'' which was used in the retrieved pair, while the baseline model chose ``precise''. A similar behavior is found with ``examen'' in the second example. This behavior helps the proposed SEG-NMT generate a translation of which style and choice of vocabulary match better with translations from a training corpus, which improves the overall consistency of the translation.

\vspace{-7pt}
\paragraph{Efficiency}
In general, there are two points at which computational complexity increases. The first point occurs at the retrieval stage which incurs almost no overhead as we rely on an efficient search engine (which retrieves a pair within several milliseconds.) In the translation stage, the complexity of indexing the key-value memory grows w.r.t. the \# of tokens in the retrieved pairs. This increase is however constant with a reasonably-set max \# of retrieved pairs. Note that the memory can be pre-populated for all the training pairs.

%\vspace{-3mm}
\section{Conclusion}
%\vspace{-3mm}

We proposed a practical, non-parametric extension of attention-based neural machine translation by utilizing an off-the-shelf, black-box search engine for quickly selecting a small subset of training translation pairs. The proposed model, called SEG-NMT, then learns to incorporate both the source- and target-side information from these retrieved pairs to improve the translation quality. We empirically showed the effectiveness of the proposed approach on the JRC-Acquis corpus using six language pair-directions. 

Although the proposed approach is in the context of machine translation, it is generally applicable to a wide array of problems. By embedding an input of any modality into a fixed vector space and using approximate search\citep{FAISS}, this approach can, for instance, be used for open-domain question answering, where the seamless fusion of multiple sources of information retrieved by a search engine is at the core. We leave these as future work.

\section*{Acknowledgments}
KC thanks support by eBay, TenCent, Facebook, Google and NVIDIA. This work was partly supported by Samsung Advanced Institute of Technology (Next Generation Deep Learning: from pattern recognition to AI). This work was also supported in part by and the HKU Artificial Intelligence to Advance Well-being and Society (AI-WiSe) Lab.

% \small
\bibliography{emnlp2017}
\bibliographystyle{aaai}

\end{document}